\documentclass[letterpaper, 10 pt, conference]{ieeeconf}  

\IEEEoverridecommandlockouts                              

\usepackage[hidelinks]{hyperref}       
\usepackage{url}            
\usepackage{booktabs}       
\usepackage{amsfonts}       
\usepackage{nicefrac}       
\usepackage{microtype}      
\usepackage{graphicx}
\usepackage{multirow}
\usepackage{wrapfig}
\usepackage{amsmath}
\usepackage{dsfont}
\usepackage{lipsum}
\usepackage{tikz}
\usepackage{subcaption}
\usepackage[font=small,labelfont=bf]{caption}
\usepackage{algorithm}
\usepackage{algorithmicx}
\usepackage[noend]{algpseudocode}
\usepackage{adjustbox}
\usepackage[mode=buildnew]{standalone}
\usepackage[
    separate-uncertainty=true,
    table-align-uncertainty=true,
    detect-mode=true]{siunitx}
\usepackage{xcolor}
\usepackage{flushend}
\usepackage{svg}
\usepackage{geometry}
\usetikzlibrary{calc}
\usetikzlibrary{arrows.meta}
\usetikzlibrary{decorations.pathreplacing}
\definecolor{custom_gray}{RGB}{83,86,90}
\definecolor{cardinal}{RGB}{140,21,21}

\definecolor{red}{HTML}{FF0000}

\newcommand{\ph}[1]{{\textbf{#1}:}}

\newcommand{\algname}{\text{DiFS}}
\newcommand{\algfullname}{\text{Diffusion-based Failure Sampling}}

\usepackage[style=ieee, url=false,doi=false,eprint=false,natbib=false,mincitenames=1,maxcitenames=1,dashed=false,isbn=false, maxbibnames=99]{biblatex}
\usepackage[capitalize]{cleveref}
\AtEveryBibitem{\clearfield{issn}}
\AtEveryCitekey{\clearfield{issn}}

\title{\LARGE \bf
Diffusion-Based Failure Sampling for Evaluating Safety-Critical Autonomous Systems
}

\author{Harrison Delecki$^{1\star}$, Marc R. Schlichting$^{1\star}$,  Mansur Arief$^{1}$, Anthony Corso$^{1}$,\\ Marcell Vazquez-Chanlatte$^{2}$,  Mykel J. Kochenderfer$^{1}$
\thanks{$^{1}$Department of Aeronautics and Astronautics, Stanford University  (e-mail: \{hdelecki, mschl, ariefm, acorso, mykel\}\!@stanford.edu).}%
\thanks{$^{2}$ Nissan Advanced Technology Center Silicon Valley, (e-mail: marcell.chanlatte@nissan-usa.com)}%
\thanks{$^{\star}$Indicates equal contribution}%
}

\begin{document}

\maketitle
\thispagestyle{empty}
\pagestyle{empty}

\begin{abstract}

Validating safety-critical autonomous systems in high-dimensional domains such as robotics presents a significant challenge. Existing black-box approaches based on Markov chain Monte Carlo may require an enormous number of samples, while methods based on importance sampling often rely on simple parametric families that may struggle to represent the distribution over failures. We propose to sample the distribution over failures using a conditional denoising diffusion model, which has shown success in complex high-dimensional problems such as robotic task planning. We iteratively train a diffusion model to produce state trajectories closer to failure. We demonstrate the effectiveness of our approach on high-dimensional robotic validation tasks, improving sample efficiency and mode coverage compared to existing black-box techniques.

\end{abstract}

\section{Introduction}

Greater levels of automation are being considered in applications such as self-driving cars \cite{badue2021selfdrivingsurvey} and air transportation systems \cite{straubinger2020overviewUAM} with the promise of improved safety and efficiency \cite{baheri2023safety}. These safety-critical systems require thorough validation for acceptance and safe deployment~\cite{Corso2021survey, corso2021transfer}. A key step in safety validation is to identify the potential failure modes that are most likely to occur.


Finding likely failures in safety-critical autonomous systems is challenging for several reasons. First, the search space is high-dimensional due to the large state spaces and long trajectories over which autonomous systems operate. Second, failures tend to be rare because systems are typically designed to be relatively safe. Direct Monte Carlo sampling may require an enormous number of samples to discover rare failures. Third, autonomous systems can exhibit multiple potential failure modes that may be difficult to uncover.

Many previous approaches frame validation as a generic optimization problem \cite{dreossi2015efficient, Akazaki2018falsification, tuncali2019rapidly}. The main disadvantage of these methods is that they tend to converge to a single failure, such as the most extreme or most likely example \cite{lee2020adaptive}. The mode-seeking behavior of optimization-based methods is insufficient to capture the potentially multimodal distribution over failure trajectories.  Markov chain Monte Carlo \cite{norden2019efficient} and adaptive importance sampling \cite{kim2016improvingAircraftCollisionCEM, zhao2017acceleratedIS, okelly2018scalableAVTestingCEM} methods have also been used to sample failure trajectories. However, existing approaches are only feasible in low-dimensional spaces, may require domain knowledge to perform well, and typically rely on simple parametric distributions. These limitations make existing techniques less useful for the validation of autonomous systems in a general setting.

\begin{figure}[!t]
    \centering
    \includesvg[width=\linewidth]{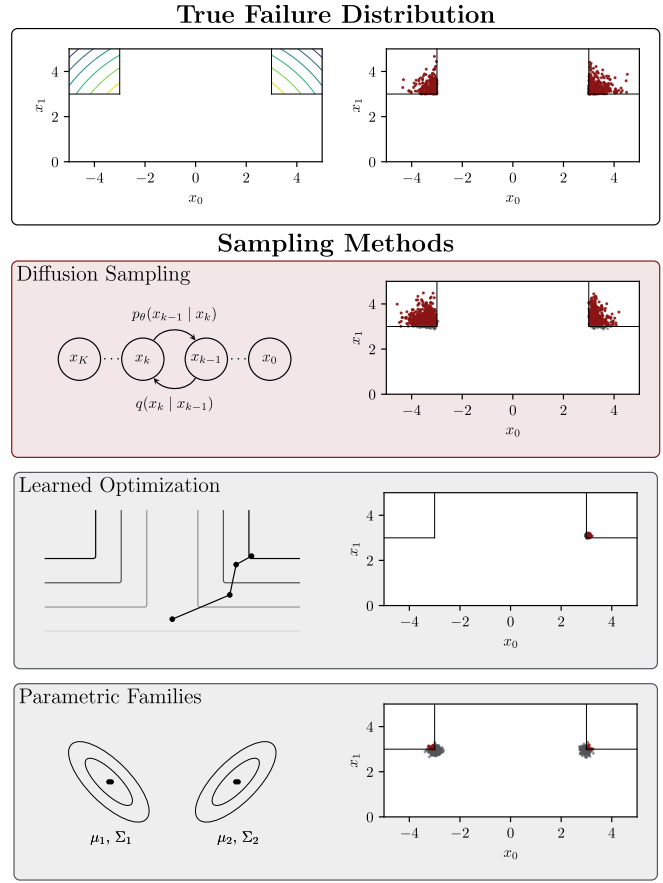}
    \caption{Illustration of a toy rare failure event sampling problem. Events occur when samples from the unit Gaussian have $|x_0|\geq 3$ and $x_1\geq3$. The top figure shows the true failure density and Monte Carlo samples. Sampling methods based on learned optimization tend to suffer mode collapse, while simple parametric families struggle to capture complex distributions. Our diffusion-based sampling approach reliably models complex multimodal failure distributions and scales to high dimensional systems.}
    \label{fig:abstract}
\end{figure}

Recently, diffusion models \cite{ho2020denoising} have demonstrated success in modeling high-dimensional distributions in diverse applications like image generation \cite{song2020score, nichol2021improved, zhang2023shiftddpms} and robotic planning~\cite{janner2022planning, wang2024diffail}. This work aims to enable efficient black-box sampling of the distribution over failure trajectories in autonomous systems. Our method generates system disturbances using a denoising diffusion model conditioned on a desired system robustness.  We propose an adaptive training algorithm that updates the proposal distribution based on a lower-quantile of the sampled data, meaning that the proposal progressively moves samples towards the failure region. We evaluate the proposed approach on five validation problems ranging from a simple two-dimensional example to a complex $1200$-dimensional aircraft ground collision avoidance autopilot. We compare our method with two baselines using metrics that assess the fidelity and diversity of failure samples. Our results show that our method outperforms baseline methods in modeling high-dimensional, multimodal failure distributions. \cref{fig:abstract} illustrates the performance of the method compared to baselines on a toy problem. Our code is available on GitHub.\footnote[1]{\href{https://github.com/sisl/DiFS}{https://github.com/sisl/DiFS}} Our contributions are as follows:
\begin{itemize}
    \itemsep0em
    \vspace{-1mm}
    \item We model the distribution over failure trajectories in autonomous systems using a robustness-conditioned diffusion model.
    \item We propose \algfullname{} (\algname{}), a training algorithm that adaptively brings a diffusion model-based proposal closer to the failure distribution.
    \item We evaluate \algname{} on five sample validation problems, demonstrating superior performance compared to baselines in terms of failure distribution fidelity, failure diversity, and sample efficiency.
\end{itemize}

\begin{figure*}[!t]
    \centering
    \begin{tikzpicture}
    \node[draw,thick,black,rounded corners,align=center,anchor=center] at (0,0) (train) {\includestandalone{figures/zoom/diffusion}\\Train Diffusion Model $p_{\theta_i}(\mathbf{x}\mid r)$};

    \node[draw,thick,black,rounded corners,align=center,anchor=center] at ($(train.east)+(6.5,-2.1)$) (sample) {\input{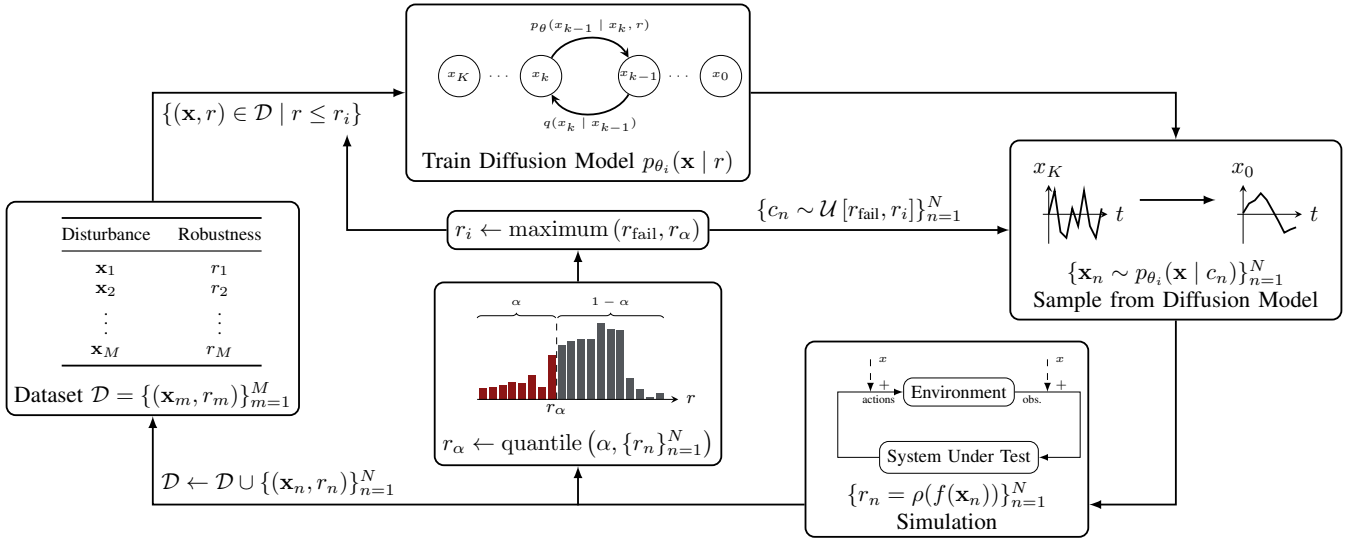}\\$\{\mathbf{x}_n\sim p_{\theta_i}(\mathbf{x}\mid c_n)\}_{n=1}^N$\\Sample from Diffusion Model};
    
    \node[draw,thick,black,rounded corners,align=center] at ($(sample)+(-3.5,-3.2)$) (eval) {\includestandalone{figures/zoom/simulation}\\$\{r_n=\rho(f(\mathbf{x}_n))\}_{n=1}^N$\\Simulation};
    \node[draw,thick,black,rounded corners,align=center,anchor=center] at (sample.west -| train.south) (ri) {$r_i\leftarrow \mathrm{maximum}\left(r_\mathrm{fail},r_\alpha\right)$};
    \node[draw,thick,black,rounded corners,align=center] at ($(ri)+(0,-2.2)$) (ralpha) {\includestandalone{figures/zoom/quantile}\\ $r_\alpha \leftarrow \mathrm{quantile}\left(\alpha,\{r_n\}_{n=1}^N\right) $};
    \node[draw,thick,black,rounded corners,align=center] at ($(ralpha)+(-6.5,1)$) (dataset) {\includestandalone{figures/zoom/table}\\Dataset $\mathcal{D} = \{(\mathbf{x}_m,r_m)\}_{m=1}^M$};

    \draw[thick, -latex] (ri.east) -- (sample.west) node[midway, above] {$\{c_n\sim \mathcal{U}\left[r_\text{fail},r_i\right]\}_{n=1}^N$};
    \draw[thick,-latex] (train.east) -| ($(sample.north)+(0,0)$) {};
    \draw[thick,-latex] ($(sample.south)+(0,0)$) |- ($(eval.east)+(0,-1)$) {};
    \draw[thick, -latex] ($(eval.west)+(0,-1)$) -| (dataset.south) node[right,pos=0.625] {$\mathcal{D}\leftarrow\mathcal{D}\cup\{(\mathbf{x}_n,r_n)\}_{n=1}^{N}$};
    \draw[thick, -latex] ($(ralpha.south) - (0,0.6)$) -- (ralpha.south);
    \draw[thick, -latex] (dataset.north) |- (train.west) node[pos=0.4,right] (selection) {$\{(\mathbf{x},r)\in \mathcal{D}\mid r\leq r_i\}$};
    \draw[thick, -latex] (ri.west) -| ($(selection.south)+(1.3,0)$) {};
    \draw[thick, -latex]  (ralpha.north -| ri) -- (ri.south) ;
    
\end{tikzpicture}
    \caption{Illustration of the proposed approach. We train a conditional denoising diffusion model to generate disturbance trajectories that are added to the simulation of a physical system either as action or observation noise. After ranking the disturbances based on their evaluated robustness in increasing order, we train the next iteration of the disturbance-generating diffusion model based on all the samples that we have previously seen whose robustness is lower than the bottom $\alpha$ quantile of samples from the $(i-1)$-th diffusion model iteration.}
    \label{fig:main}
\end{figure*}

\section{Related Work}
Traditional validation methods often assume access to a complete mathematical model of the system under test \cite{clarke2018handbookmodelchecking}. These methods scale poorly to large, complex systems. Black-box approaches tend to scale well, since they only require the ability to simulate the system under test.

Previous work in black-box validation generally frames the problem as optimization, where the objective is to find an input disturbance to the system that leads to system failure. Black-box approaches rely on classical gradient-free optimization \cite{dreossi2015efficient, Akazaki2018falsification} and path planning \cite{tuncali2019rapidly}. An approach called adaptive stress testing (AST) \cite{Lee2019a} outputs disturbances that are likely to lead to a system failure. The main drawback of these algorithms is that they tend to converge to a single failure mode. See the survey by \textcite{Corso2021survey} for more on black-box safety validation algorithms.

There are parametric and non-parametric approaches to sampling the failure distribution. Parametric methods such as the cross-entropy method (CEM) \cite{kim2016improvingAircraftCollisionCEM, okelly2018scalableAVTestingCEM} iteratively update a parametric family towards the failure distribution. However, the choice of parametric family may limit performance in complex, high-dimensional distributions. Non-parametric methods like adaptive multilevel splitting (AMS) \cite{cerou2007adaptive, norden2019efficient} use Markov chain Monte Carlo (MCMC) to iteratively bring a set of particles towards the failure distribution. However, black-box MCMC-based methods struggle to scale to high-dimensional validation problems \cite{delecki2023model}. Our aim is to address these limitations by training a diffusion model to sample the failure distribution.

Recent validation methods have found success by incorporating generative models in various safety validation algorithms. For example, normalizing flows have been used to perform more efficient gradient-based sampling of the failure distribution \cite{sinha2020neural}. A different approach fits a normalizing flow to a smooth approximation of the distribution over failures and draws samples from the trained flow \cite{gibson2023flowgenmodels}. The drawback of these methods is that both require a differentiable simulation, which may not be available. Generative adversarial networks have also been used to falsify signal-temporal logic expressions \cite{peltomaki2023requirementgans}. This technique targets single failure modes, and does not account for the distribution over failures.

Diffusion models are a class of generative model that generate samples by learning to reverse an iterative data noising process \cite{ho2020denoising, song2020score}. These models have shown success in modeling high-dimensional, multimodal distributions in applications such as image generation and robot motion planning \cite{janner2022planning, carvalho2023motion, zhang2023shiftddpms, wang2024diffail}. Based on these results, we explore using diffusion models to learn a distribution over disturbances that lead to failure in safety-critical autonomous systems.

\section{Methods}
In this section, we present our method to approximately sample from the distribution over failures in safety-critical systems, \algfullname{} (\algname{}). First, we introduce necessary notation and formulate the distribution over failures. Next, we describe how diffusion models may be used for validation given sufficient failure data. Finally, we describe \algname{}, which collects failure samples during training.

\subsection{Problem Formulation}
Consider a system under test that takes actions in an environment after receiving observations. We denote a system state trajectory as $\mathbf{s}=[s_1,\dots,s_t]$ where $s_t$ is the state of the environment at time $t$. We also define a robustness property $\rho(\mathbf{s})$ over a state trajectory. We define a system failure to occur when the robustness is below a given threshold, $\rho (\mathbf{s}) \leq r_{\text{fail}}$.

We perturb the environment with disturbance trajectories $\mathbf{x}$ to induce the system under test to violate the robustness constraint. For example, sensor noise and stochastic dynamics are potential disturbances that could lead to failure. We denote the simulation of the system under test with environment disturbances by a dynamics function $f$. Note that $f$ represents the dynamics of both the system under test and the environment. We assume that disturbances determine all sources of stochasticity in the environment. Therefore, the resulting state trajectory under disturbances is written as $\mathbf{s} = f(\mathbf{x})$.

A disturbance trajectory $\mathbf{x}$ has probability density $p(\mathbf{x})$, reflecting how likely we expect disturbances to be in the environment. For example, we might expect smaller disturbances to be more likely in general. Our goal is to sample disturbance trajectories that violate the robustness constraint, or sample from the conditional distribution 
\begin{equation}
    \label{eq:exact-failure-dist}
    p^{\star}(\mathbf{x} \mid r_{\text{fail}}) \propto \mathds{1} [\rho(f(\mathbf{x})) \leq r_{\text{fail}}] p(\mathbf{x})
\end{equation}
where $\mathds{1}\left[\cdot\right]$ is the indicator function.

Representing the distribution $p^{\star}(\mathbf{x} \mid r_{\text{fail}})$ analytically is typically intractable even for simple autonomous systems. Sampling from this distribution is often difficult due to the high dimensionality, rareness, and multimodality of failure trajectories in autonomous systems. We propose to sample from this distribution by training a denoising diffusion model to generate disturbance trajectories from the failure distribution. Diffusion models are particularly appealing for this application due to their success in modelling complex high-dimensional data in domains like robotic planning, their stability during training, and their potential to compute likelihoods. In the following subsection, we present a brief discussion of diffusion models in the context of our Failure sampling problem.

\subsection{Diffusion for Validation}
We use a denoising diffusion probabilistic model to represent the failure distribution \cite{ho2020denoising, nichol2021improved}. Diffusion models progressively add noise to data, and learn to reverse this process to generate new samples. Let $\mathbf{x}_k$ be the disturbance trajectory at the $k$-th diffusion step, with $k=0$ corresponding to a disturbance trajectory from the target data distribution. The forward diffusion process starting from data $\mathbf{x}_0$ is
\begin{equation*}
q\left(\mathbf{x}_{1: K} \mid \mathbf{x}_0\right)=\prod_{k=1}^K \mathcal{N}\left(\mathbf{x}_k \mid \sqrt{1-\beta_k} \mathbf{x}_{k-1}, \beta_k \mathbf{I}\right)
\end{equation*}
where $\beta_1, \ldots, \beta_K$ is a predetermined variance schedule controlling the noise magnitude at each diffusion step. As the noise accumulates, the data approximately transforms into a unit Gaussian at the final diffusion step. 




To generate disturbances with a desired level of robustness, we train a conditional diffusion model. This model is trained on a dataset $\mathcal{D} = \{(\mathbf{x}_m, r_m)\}_{m=1}^M$ containing disturbance trajectories paired with corresponding robustness level. Starting from sampled noise, the diffusion model gradually denoises the sample at each step. This reverse process is
\begin{equation*}
p_\theta\left(\mathbf{x}_{0: K} \mid r\right)=p\left(\mathbf{x}_K\right) \prod_{k=1}^K \mathcal{N}\left(\mathbf{x}_{k-1} \mid \boldsymbol{\mu}_\theta\left(\mathbf{x}_k, k, r\right), 
 \beta_k \mathbf{I}\right)
\end{equation*}
where $p\left(\mathbf{x}_K\right)=\mathcal{N}(\mathbf{x}_K \mid \mathbf{0}, \mathbf{I})$ and $\theta$ represents the parameters of the diffusion model. In the appendix, we perform an ablation that shows that conditioning the model significantly improves generalization compared to an unconditional model. Our goal is to train a conditional diffusion model $p_{\theta}(\mathbf{x} \mid r)$ to sample the failure distribution in \cref{eq:exact-failure-dist}. If failure events are rare under $p(\mathbf{x})$, it may be very difficult to acquire sufficient training data.

\subsection{Adaptive Training}
 Inspired by adaptive methods such as CEM and AMS, we propose an iterative algorithm that adaptively selects the robustness threshold $r_i$ during training. The algorithm has two key features. First, the algorithm collects information about the conditional distribution by sampling trajectories with robustness levels from the current $r_i$ to the target. Next, we assess how good the model is by evaluating the true robustness associated with each sample, and adaptively update the target robustness threshold. These features balance between collecting information about the target distribution and improving model performance.

Initially, the diffusion model is trained on samples from $p(\mathbf{x})$ so that the distribution is close to the disturbance model. At each iteration, the algorithm performs four main steps consisting of sampling disturbances, evaluating their robustness level, updating the robustness threshold, and training the diffusion model.


To sample disturbances from the model, we need to choose an input desired robustness level. One option would be to use the failure robustness level $r_{\text{fail}}$. However, this may lead to poor performance if there is limited data near $r_{\text{fail}}$. Instead, we sample conditioning robustness values uniformly between the current robustness threshold and the target, $c_n \sim \mathcal{U}\left[r_{\text{fail}},r_i\right]$. By uniformly sampling the input conditions, we aim to balance generating samples close to failure when the model can accurately represent the target distribution with collecting new disturbance data in unexplored regions. Disturbances are sampled from the diffusion model according to $\mathbf{x}_n \sim p_{\theta_i}(\mathbf{x} \mid c_n)$, and we evaluate the robustness $r_n$ of each sampled $\mathbf{x}_n$. We compute the updated robustness threshold $r_{i+1}$ according to the bottom $\alpha$ quantile of sampled robustness. This ensures that a minimum fraction $\alpha$ of the samples have at most a robustness level $r_{i+1}$.

Since our objective is to model a distribution conditioned on the desired robustness level, we maintain a running dataset of collected samples rather than only training on the newly sampled data as in CEM. After updating the robustness threshold, we append all samples to the dataset. For training the next iteration of the diffusion model, we only use samples with a robustness of $r\leq r_{i}$. We do this to focus the training effort towards the areas of decreasing robustness that we ultimately wish to sample from. Finally, we train the diffusion model on this updated dataset. We call this algorithm \algfullname{}, (\algname{}). \algname{} as described here is outlined in \cref{alg:diffusion-iterative-sample}.

\begin{algorithm}[H]
    \footnotesize
    \caption{Diffusion-based Failure Sampling (DiFS)}
    \label{alg:diffusion-iterative-sample}
    \begin{algorithmic}[1]
        \Require simulator $f(\cdot)$, robustness $\rho(\cdot)$, failure robustness $r_\mathrm{fail}$
        \Require initial denoising diffusion model $p_{\theta_0}(\mathbf{x} \mid r)$
        \Require prior disturbance model $p(\mathbf{x})$, quantile threshold $\alpha$
        \Function{${\textproc{DiFS}}(f, \rho, r_\mathrm{fail}, \alpha, p_{\theta}(\mathbf{x} \mid r), p(\mathbf{x}))$}{}
        \State $i \leftarrow 0$
            \State sample $\{\mathbf{x}_n\sim p(\mathbf{x})\}_{n=1}^N$
            \State evaluate robustness $\{r\leftarrow\rho(f(\mathbf{x}_n))\}_{n=1}^N$
            \State compute $r_\alpha\leftarrow\mathrm{quantile}(\alpha,\{r_n\}_{n=1}^N)$
            \State compute $r_i\leftarrow\mathrm{maximum}(r_\mathrm{fail},r_\alpha)$
            \State $\mathcal{D} \leftarrow \{(\mathbf{x}_n, r_n)\}_{n=1}^N$
            \State $\theta_i \leftarrow \textproc{Train}(\theta_0, \mathcal{D})$
            \While {$r_i > r_{\text{fail}}$}
                \State sample robustness conditions $\{c_n \sim \mathcal{U}\left[r_{\text{fail}},r_i\right])\}_{j=1}^N$
                \State sample disturbances $\{\mathbf{x}_n \sim p_{\theta_i}(\mathbf{x} \mid c_n)\}_{n=1}^N$ 
                \State evaluate robustness $\{r\leftarrow\rho(f(\mathbf{x}_n))\}_{n=1}^N$
                \State compute $r_\alpha\leftarrow\mathrm{quantile}(\alpha,\{r_n\}_{n=1}^N)$
                \State compute $r_{i+1}\leftarrow\mathrm{maximum}(r_\mathrm{fail},r_\alpha)$
                \State $\mathcal{D} \leftarrow \mathcal{D} \cup \{(\mathbf{x}_n, r_n)\}_{n=1}^N$
                \State $\theta_{i+1} \leftarrow \textproc{Train}(\theta_i, \{(\mathbf{x},r)\in\mathcal{D}\mid r\leq r_i\})$
                \State $i\leftarrow i+1$

            \EndWhile
            \State \Return $\theta_{i}$
        \EndFunction
    \end{algorithmic}
\end{algorithm}

\section{Experiments and Results}
This section covers the experimental setup used to evaluate the proposed approach. First, we discuss validation problems. Next, we discuss baselines, metrics, qualitative evaluation of multimodality, and experimental setup.


\subsection{Validation Problems}

We demonstrate the feasibility of our diffusion-model based validation framework using a two-dimensional toy problem and four robotics validation problems. 

\ph{2D Toy}
The toy problem is adapted from~\textcite{sinha2020neural}. Disturbances are sampled from a unit 2D Gaussian, and robustness is defined as $\rho(\mathbf{x})=3.0 - \mathrm{minimum}(\lvert x_0\rvert,x_1)$. Failure occurs when $\rho(\mathbf{x}) \leq 0$. This problem has two failure modes, corresponding to the regions where $(x_0 \leq -3, x_1 \geq 3)$ and where $(x_0 \geq 3, x_1 \geq 3)$.

\ph{Pendulum}
We consider an underactuated, inverted pendulum environment from Gymnasium\footnote{\url{https://github.com/Farama-Foundation/Gymnasium}} that is stabilized by a PD-controller subject to input torque disturbances. We assume that the torque perturbations at each time step are distributed according to $\mathcal{N}(\mathbf{0}, 0.5\mathbf{I})$, and limit the trajectories to $100$ time steps. Trajectory robustness is defined as the maximum absolute deflection of the pendulum from vertical, and we define failures to occur when the absolute deflection exceeds $30\deg$ from vertical. The system has two failure modes where the pendulum can fall left or right.

\ph{Intersection}
We also consider an autonomous driving validation problem based on the intersection environment of HighwayEnv.\footnote{\url{https://github.com/Farama-Foundation/HighwayEnv}} The system under test is a rule-based policy designed to navigate a four-way intersection. We consider a scenario with two vehicles: the ego vehicle which follows the rule-based policy and an intruder vehicle which uses the intelligent driver model \cite{treiber2000IDM}. The ego vehicle traverses the intersection straight while the intruder vehicle makes a turn onto ego vehicle's lane. The ego vehicle's rule-based policy changes the commanded velocity based on the time to collision with the intruder vehicle. We perturb the observed position and velocity components of the intruder vehicle over 24 timesteps, resulting in a 96-dimensional problem. A failure is defined if the distance between ego and intruder vehicle is less than $0.06$ units of length.

\ph{Lunar Lander}
We validate a heuristic control policy for the continuous LunarLander environment from Gymnasium. We add noise to the lander's obervations of its horizontal position and orientation over $200$ time steps, giving this problem $400$ dimensions. Disturbances for position and orientation are zero-mean Gaussians with a variance of $0.31$. We define failure to occur when the lander touches down outside of the landing pad. The trajectory robustness corresponds to the distance to the edge of the pad at landing.

\ph{F-16 Ground Collision Avoidance} Finally, we validate an automated ground collision avoidance system (GCAS) for the F-16 aircraft. The GCAS detects impending ground collisions and executes a series of maneuvers to prevent collision. We adapt the flight and control models from an open source verification benchmark \cite{heidlauf2018verification}. We add disturbances to observations of orientation and angular velocity. The aircraft starts at an altitude of $\SI{600}{ft}$, pitched down $\SI{30}{deg}$, and is rolled $\SI{20}{deg}$. The controller first rolls the wings level and then pitches up to avoid collision. We noise the $6$ observations for $200$ time steps, giving this problem $1200$ dimensions.

\begin{table}[!t]
\centering
\caption{Training hyperparameters for \algname{}.}
\begin{adjustbox}{max width=\columnwidth}
\begin{tabular}{@{}lrrrrr@{}}
\toprule
Hyperparameter             & Toy & Pendulum & Intersection & Lander & F-16 \\ \midrule
Sample budget     &      \num{50000}      &       \num{50000}    & \num{100000} & \num{100000} & \num{100000}  \\
Samples $/$ iter. &      \num{10000}      &       \num{10000}    & \num{10000} & \num{10000} & \num{10000}  \\
Train steps $/$ iter. &      \num{10000}      &   \num{10000}   & \num{10000}& \num{20000} & \num{20000}         \\
$\alpha$              &      \num{0.5}      &        \num{0.5}       &      \num{0.5}     & \num{0.5} & \num{0.3}   \\
\bottomrule
\end{tabular}
\end{adjustbox}
\label{tab:hyperparameters}
\end{table}

\begin{table}[!t]
\centering
\caption{Results of Monte Carlo sampling for each problem.}
\begin{adjustbox}{max width=\columnwidth}
\begin{tabular}{@{}lrrrrr@{}}
\toprule
             & Toy & Pendulum & Intersection & Lander & F-16 \\ \midrule
Samples     &      \num{1e7}      &       \num{2e7}    & \num{6e6} & \num{2e9} & \num{1e7}  \\
Failure Probability Estimate &      \num{3.5e-5}      &       \num{2.1e-4}    & \num{2.0e-4} & \num{5.6e-7} & \num{1.1e-4}  \\
Sampling Time (days) &      \num{0.0}      &   \num{3.5}   & \num{4.2}& \num{520} & \num{4.7}         \\
\bottomrule
\end{tabular}
\end{adjustbox}
\vspace{-5mm}
\label{tab:mc-results}
\end{table}

\subsection{Baselines}

We compare \algname{} against two baseline validation techniques. First, we consider the cross-entropy method with a two-component Gaussian mixture model proposal (CEM-$2$). CEM-$2$ iteratively updates the parametric proposal by minimizing the KL divergence to the failure distribution \cite{rubinstein2004CEM}. Finally, we also consider adaptive stress testing (AST), which is a validation algorithm for sequential systems that frames validation as a Markov decision process (MDP) \cite{lee2020adaptive}. At each time step, the agent observes the system state and chooses an input disturbance. We solve the MDP using the proximal policy optimization reinforcement learning algorithm implemented in the stable-baselines library.\footnote{\url{https://github.com/DLR-RM/stable-baselines3}}

\subsection{Metrics}
We quantitatively evaluate the samples of state trajectories generated by \algname{} and the baselines after training using failure sample density, coverage, and failure rate metrics. The density and coverage metrics of \textcite{naeem2020fidelitydiversity} evaluate failure sample similarity to the true failure distribution, while the failure rate evaluates sample efficiency.

\ph{Density} The sample density metric, denoted $D_s$, measures the fraction of ground-truth sample neighborhoods that include the generated samples. Neighborhoods are defined by the $k$-nearest neighbors. Density is unbounded and may be greater than $1$ if the true samples are densely clustered around the generated samples.

\ph{Coverage}  Coverage, denoted by $C_s$, measures the fraction of true samples whose neighborhoods contain at least one generated sample. Coverage is bounded between $0$ and $1$, with $1$ being perfect coverage. 

\ph{Failure rate} The failure rate $R_{\text{fail}}$ is the proportion of samples drawn after result in failure after training. While density and coverage measure how well the sampled failures match the true distribution, it is also important to avoid wasting time evaluating safe samples. A failure rate of $1$ indicates a sample-efficient generation of failures.

\begin{table}[t!]
\sisetup{
         table-text-alignment=right,
         detect-weight=true,
        }
    \begin{center}
    \caption{Results comparing \algname{}, CEM-2, and AST across various problems. Higher is better for all metrics.}
    \begin{adjustbox}{max width=\columnwidth}
        \begin{tabular}{@{}l l
        S[table-format=3.2(1), round-precision=4, separate-uncertainty=true, table-align-uncertainty]
        S[table-format=3.1, round-precision=1, table-align-uncertainty]
        S[table-format=1.2e-1, scientific-notation = false, table-align-uncertainty]
        S[table-format=4.1, round-precision=1, table-align-uncertainty] 
        @{}}
        \toprule
         & $\text{Method}$ & $\text{$D_\mathbf{s}$} \uparrow$ & $\text{$C_\mathbf{s}$} \uparrow$ & $R_{\text{fail}} \uparrow$ & $\text{Time (hr)}$ \\ 
        \midrule
        
        \multirow{3}{*}{\parbox{1.5cm}{Toy}} & \textbf{\algname{}} & \bfseries \num{0.998(12)} & \bfseries\num{0.980(20)} & \bfseries \num{0.896(47)} & \num{0.61} \\ 
        & CEM-2 & \num{0.636(371)} & \num{0.257(150)} & \num{0.182(251)} & \num{0.01} \\ 
        & AST & \num{0.501(462)} & \num{0.108(67)} & \num{0.868(107)} & \num{0.13} \\ 
        
        \midrule
        
        \multirow{3}{*}{\parbox{1.5cm}{Pendulum}} & \textbf{\algname{}} & \bfseries \num{0.872(68)} & \bfseries \num{0.901(42)} & \bfseries  \num{0.795(91)} & \num{2.93} \\ 
        & CEM-2 & \num{0.453(134)} & \num{0.473(4)} & \num{0.599(36)} & \num{0.22} \\ 
        & AST & \num{0.578(1)}  & \num{0.116(21)} & \num{0.337(78)} & \num{2.58} \\ 

        \midrule
        
        \multirow{3}{*}{\parbox{1.5cm}{Intersection}} & \textbf{\algname{}} &  \bfseries\num{0.907(117)} & \bfseries \num{0.879(32)} &  \num{0.568(2)} & \num{20.19} \\ 
        & CEM-2 & \num{0.746(204)} & \num{0.064(26)} & \bfseries \num{0.877(58)} & \num{3.48} \\ 
        & AST & \num{0.367(24)}  & \num{0.049(11)} & \num{0.851(86)} & \num{26.83} \\ 

        \midrule
        
        \multirow{3}{*}{\parbox{1.5cm}{Lander}} & \textbf{\algname{}} & \bfseries \num{0.653(65)} & \bfseries \num{0.876(43)} &  \num{0.667(71)} & \num{7.24} \\ 
        & CEM-2 & $\text{---}$  & $\text{---}$ & \num{0.0} & \num{0.74} \\ 
        & AST & \num{0.113(82)}  & \num{0.031(16)} & \bfseries \num{0.912(65)} & \num{9.22} \\ 

        \midrule
        
        \multirow{3}{*}{\parbox{1.5cm}{F-$16$}} & \textbf{\algname{}} & \bfseries \num{0.886(31)} & \bfseries \num{0.981(5)} &  \bfseries \num{0.687(35)} & \num{7.59} \\ 
        & CEM-2 & \num{0.742(361)} & \num{0.341(359)} & \num{0.424(411)} & \num{0.66} \\ 
        & AST & \num{0.156(302)}  & \num{0.002(3)} & \num{0.410(489)} & \num{12.85} \\ 
        
        \bottomrule
        \end{tabular}
        \end{adjustbox}
        \label{tab:results}
    \end{center}
    \vspace{-5mm}
\end{table}

\subsection{Multimodality of Disturbances}
To demonstrate that our framework captures multimodal disturbance trajectories, we apply principal component analysis (PCA) and visually inspect potential clusters in a $2$D projection. Since not all validation problems exhibit multimodal failure disturbances, we assess unimodality using uniform manifold approximation and projection (UMAP) \cite{mcinnes2018umap}, a state-of-the-art nonlinear dimensionality reduction technique. While UMAP provides a powerful nonlinear representation, we rely on PCA to enhance interpretability of multimodality.





\subsection{Experimental Procedure}
All experiments were performed on a machine with a single GPU with $11$ GB of GPU memory. Training hyperparameters used for \algname{} are shown in \cref{tab:hyperparameters}. Due to computational constraints, a full hyperparameter search is infeasible. However, empirically we find that experiments with the more challenging environments (i.e., lander and F-16) benefit from from a larger sample budget and smaller $\alpha$. All methods use the same number of simulations across $5$ random seeds.

Density and coverage metrics require ground-truth samples. For each problem, we collect $1000$ ground-truth samples from the failure distribution using MC sampling. The MC sampling results for each problem are shown in \cref{tab:mc-results}. Since samples were collected using various parallel clusters, we normalize sampling times to a single \SI{4.5}{\giga\hertz} core for consistency.

We compute density and coverage metrics by comparing the $1000$ ground-truth failure samples from MC to $1000$ samples from each method after training. \textcite{naeem2020fidelitydiversity} suggest a method for selecting the number of nearest neighbors using two groups of ground-truth samples. Using this method, we select $k=5$. The failure rate after training $R_\mathrm{fail}$ is evaluated using $1000$ samples.


\section{Results and Discussion}

\begin{figure*}[!t]
    \begin{subfigure}[b]{\textwidth}
        \centering
        \includegraphics[width=\linewidth]{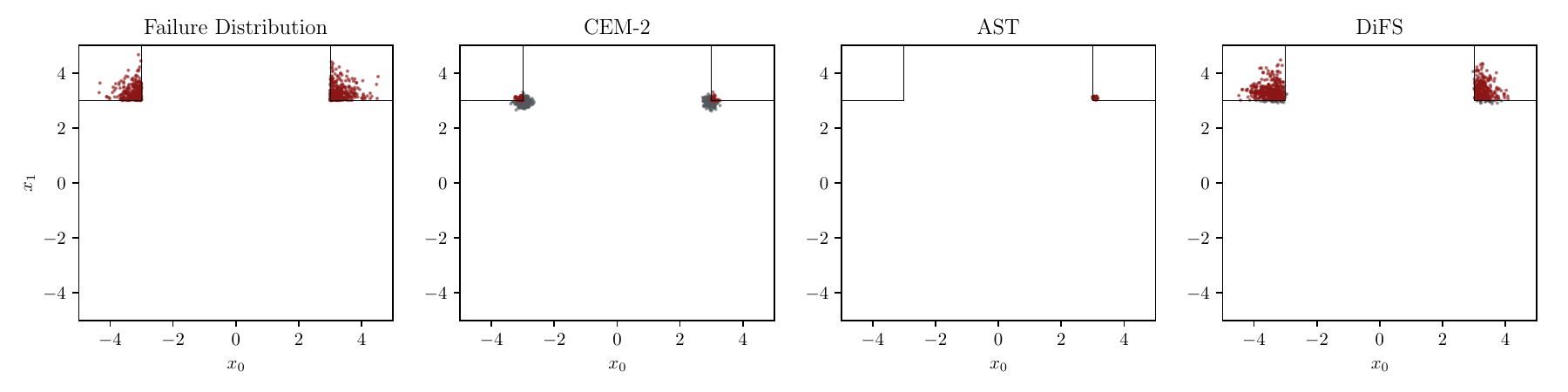}\label{fig:comparison-synthetic}
        \caption{Samples from the true failure distribution and each method for the  toy $2$D corners problem. The failure region is outlined in black.}
    \end{subfigure}
    \begin{subfigure}[b]{\textwidth}
        \centering
        \includegraphics[width=\linewidth]{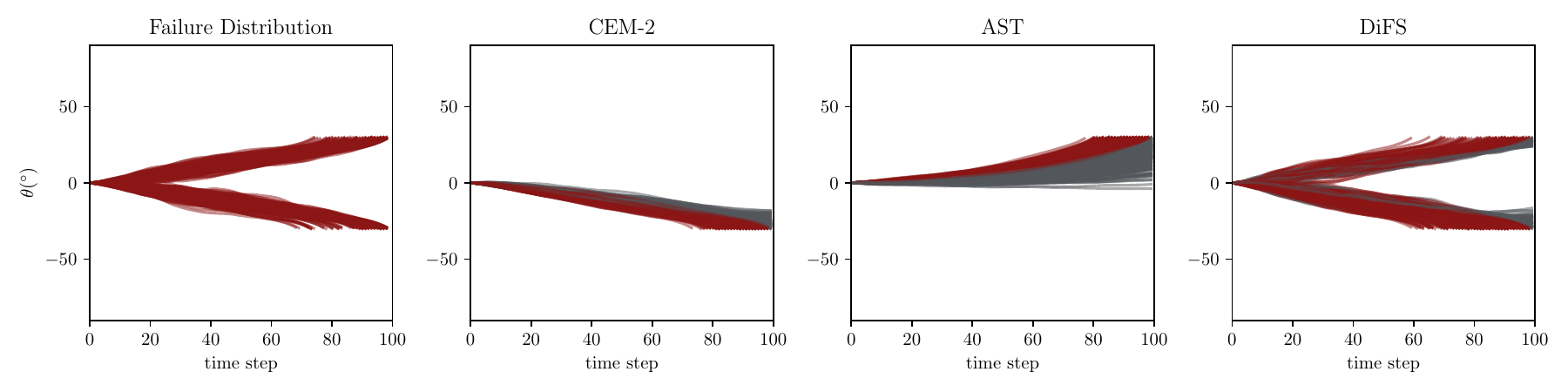}\label{fig:comparison-pendulum}
        \caption{Inverted pendulum deflection angle $\theta$ trajectories sampled from the true failure distribution and sampled from each method after training. Failure occurs when the deflection angle exceeds $\SI{30}{\deg}$.}
    \end{subfigure}
    \begin{subfigure}[b]{\textwidth}
        \centering
        \includegraphics[width=\linewidth]{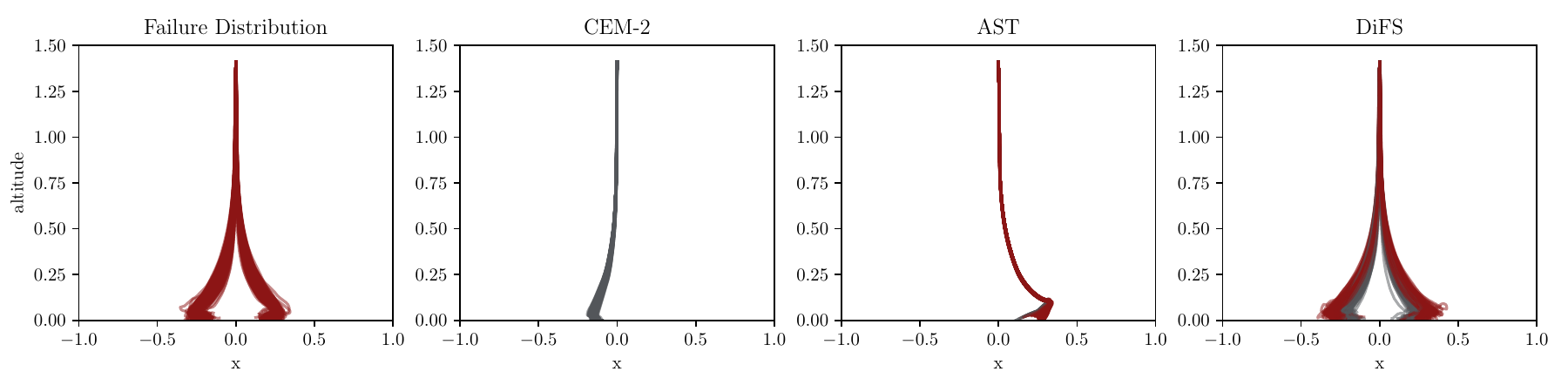}\label{fig:comparison-lander}
        \caption{Lunar lander trajectories of altitude and horizontal position sampled from the true failure distribution and sampled from each method after training. Failure occurs when the absolute value of the horizontal position at landing exceeds $0.25$.}
    \end{subfigure}
    \caption{Samples from the ground truth failure distribution and trained methods on the toy, inverted pendulum, and lunar lander problems. Ground truth samples were collected using a long run of Monte Carlo sampling. Failures are shown in red, while safe trajectories are shown in gray. In each problem, our \algname{} algorithm captures multimodal failures better than the baselines.}
    \label{fig:results}
\end{figure*}

\begin{figure*}[!t]
    \begin{subfigure}[b]{\textwidth}
        \centering
        \includegraphics[width=\linewidth]{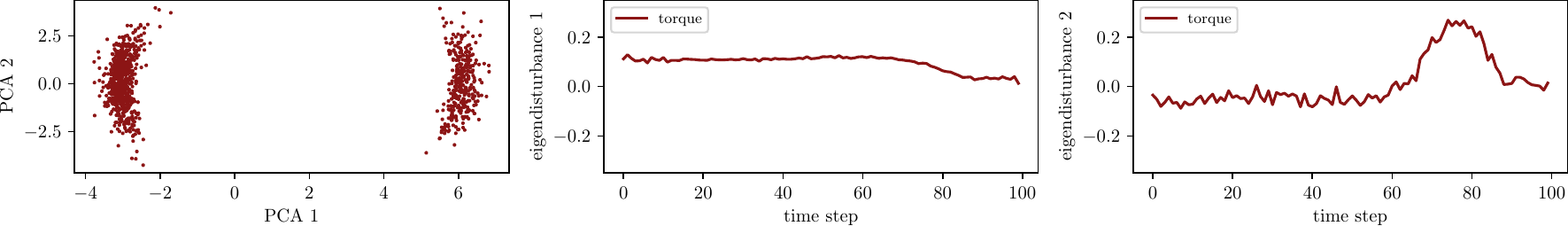}\label{fig:pca_pendulum}
        \caption{Inverted pendulum.}
         \vspace{3mm}
    \end{subfigure}
   
    \begin{subfigure}[b]{\textwidth}
        \centering
        \includegraphics[width=\linewidth]{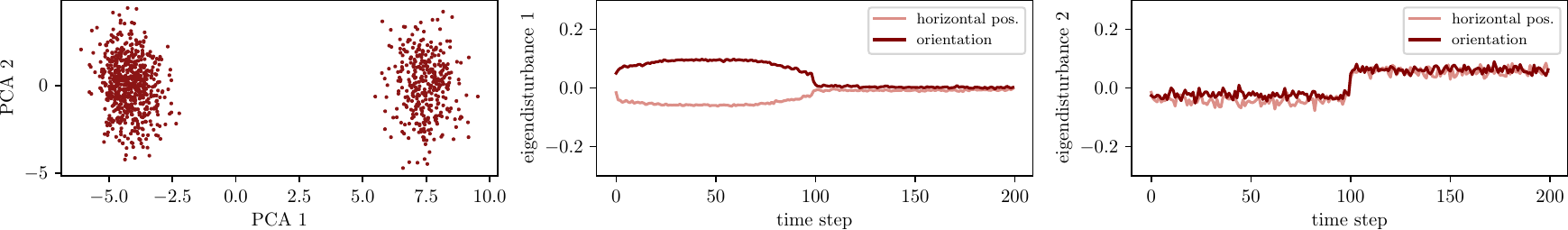}\label{fig:pca_lander}
        \caption{Lunar lander.}
    \end{subfigure}
    \caption{PCA projection of DiFS disturbances reveals multimodal behavior along the first principal component for both the inverted pendulum and the lunar lander. We also illustrate the eigenvectors corresponding to the principal components, which we refer to as eigendisturbances.}
    \label{fig:pca_results}
\end{figure*}


Quantitative results for all methods are shown in \cref{tab:results}. \algname{} generally outperforms baselines in failure rate, and also achieves a lower variance. The failure rate of \algname{} is greater than $0.66$ even on high-dimensional problems with low failure probability such as the lander and F-$16$. This high sample efficiency is due to the expressivity of the diffusion model, especially compared to the simple Gaussian proposal of CEM. AST also achieves a high failure rate for problems like the lunar lander because it optimizes a policy to sample a single most likely failure event. However, this generally leads to poor approximation of the failure distribution, which can be seen in lower density and coverage metric across all experiments when compared to DiFS.

\algname{} outperforms baselines in terms of density and coverage metrics on all problems, indicating that \algname{} better captures the true failure distribution. The high density values indicate that the \algname{} samples have a higher fidelity than the baselines. The coverage achieved by \algname{} is consistently higher than the baseline methods, indicating that \algname{} reliably discovers multimodal failure distributions. CEM and AST have coverage below $0.5$ on multimodal problems like the toy, pendulum, and lander, indicating that these baselines are prone to mode collapse.

We also report the mean wall-clock time for each method in \cref{tab:results}. CEM exhibits the fastest performance due to its efficient training process, which primarily involves fitting a GMM. In contrast, both \algname{} and AST typically require hours of computation time since they involve training machine learning models. However, the increased runtime of \algname{} comes with a large improvement in failure sampling. 


Sample trajectories drawn using each method for three problems are depicted in \cref{fig:results} where the red markers indicate failures and gray indicates safe trajectories. Comparing the results of DiFS to CEM and AST qualitatively using \cref{fig:results}, it can be seen that DiFS captures the true failure distribution the best, further supporting the claim that the diffusion model captures a distribution close to the true failure distribution. 

While DiFS outperforms the baselines on all problems, the greatest differences can be seen for high-dimensional problems with low failure probabilities. Training the diffusion model for DiFS greatly benefits from GPU availability, while evaluating costly environments in parallel requires multiple CPUs. For the more computationally expensive problems like the intersection, lander, and F-$16$, both training stages require similar wall-times. DiFS is therefore especially useful on systems with GPU and CPU resources.

Based on PCA and UMAP projections, we only identified multimodal disturbances in the inverted pendulum and lunar lander problems. We show a PCA analysis of the disturbance trajectories sampled from DiFS for the inverted pendulum and lunar lander in \cref{fig:pca_results}. For each problem, we show $2$D projections of the disturbance trajectories and visualize the principal components, or \textit{eigendisturbances}. For the inverted pendulum, projections of the $100$D disturbance trajectories show two distinct modes. Analysis of the principal components suggest these modes correspond to a positive or negative constant torque disturbance. Similarly, for the lunar lander, projections of the $400$D disturbances show two distinct modes. The first eigendisturbance suggests that the modes correspond to constant offsets in observed position and orientation, causing the lander to drift.

\section{Conclusion}
Sampling the distribution over failure trajectories in autonomous systems is key to the safe and confident deployment in safety-critical domains. This work proposes \algname{}, a method that adaptively trains a conditional denoising diffusion model to sample from the failure distribution. Experiments on five example validation problems up to $1200$ dimensions show that samples using \algname{} match the true failure distribution more accurately than baseline methods. In future work, we are interested in distilling the distribution learned by \algname{} into an online model to augment robust planning algorithms. 





\section*{Appendix}
To assess the benefit of robustness conditioning in \algname{}, we perform an ablation on the toy experiment without conditioning. We compare the number of iterations to convergence using five random seeds. Results shown in \cref{tab:cond-ablation} indicate that DiFS with conditioning converges in half as many iterations and slightly improves our evaluation metrics, suggesting improved generalization. It is also important to note that even if learning the conditional distribution is still difficult, \algname{} still performs well.

\begin{table}[!h]
\sisetup{
         table-text-alignment=right,
         detect-weight=true,
        }
\centering
\caption{Performance of \algname{} with and without robustness conditioning on the toy problem.}
\begin{adjustbox}{max width=\columnwidth}
\begin{tabular}{@{}l
S[table-format=3.2(1), round-precision=1, separate-uncertainty=true, table-align-uncertainty]
S[table-format=3.1, round-precision=1, table-align-uncertainty]
S[table-format=1.2e-1, scientific-notation = false, table-align-uncertainty]
S[table-format=3.1, round-precision=1, table-align-uncertainty]
@{}}
\toprule
 &  $\text{\# Iters.} \downarrow$ & ${D}_{\mathbf{s}} \uparrow$ & $C_{\mathbf{s}} \uparrow$ & $R_{\mathrm{fail}} \uparrow$\\ \midrule
DiFS     &     \bfseries\num{4.0(5)}  & \bfseries \num{0.998(12)} & \bfseries \num{0.839(60)} &  \bfseries  \num{0.896(47)}\\
w/o condition &   \num{8.0(20)}  & \num{0.950(15)} & \num{0.796(78)} &  \num{0.826(88)}\\
\bottomrule
\end{tabular}
\end{adjustbox}
\vspace{0.5em}
\label{tab:cond-ablation}
\vspace{-5mm}
\end{table}



\section*{Acknowledgment}
This research was supported by the National Science Foundation (NSF) Graduate Research Fellowship under Grant No. DGE-2146755. Any opinions, findings, conclusions or recommendations expressed in this material do not necessarily reflect the views of the NSF. This work was also supported by Nissan Advanced Technology Center Silicon Valley.


\printbibliography

@String { aaai        = {AAAI Conference on Artificial Intelligence (AAAI)} }

@String { aiaa_is     = {AIAA Scitech Intelligent Systems Conference (IS)} }

@String { fm          = {International Symposium on Formal Methods (FM)} }

@String { iclr        = {International Conference on Learning Representations} }

@String { icml        = {International Conference on Machine Learning (ICML)} }

@String { ieeetits    = {IEEE Transactions on Intelligent Transportation Systems} }

@String { iros        = {IEEE/RSJ International Conference on Intelligent Robots and Systems (IROS)} }

@String { itsc        = {IEEE International Conference on Intelligent Transportation Systems (ITSC)} }

@String { jair        = {Journal of Artificial Intelligence Research} }

@String { nfm        =  {NASA Formal Methods Symposium (NFM)} }

@String { neurips     = {Advances in Neural Information Processing Systems (NeurIPS)} }

@String {arxiv        = {arXiv} }

@Article{Corso2021survey,
author = {Anthony Corso and Robert J Moss and Mark Koren and Ritchie Lee and Mykel J Kochenderfer},
journal = jair,
title = {A survey of algorithms for black-box safety validation of cyber-physical systems},
year = {2021},
pages = {377--428},
volume = {72},
}

@inproceedings{Lee2019a,
    author={Lee, Ritchie and Mengshoel, Ole J and Agogino, Adrian K and Giannakopoulou, Dimitra and Kochenderfer, Mykel J},
    title={Adaptive Stress Testing of Trajectory Planning Systems},
    booktitle=aiaa_is,
    year={2019},
}

@book{clarke2018handbookmodelchecking,
  title={Handbook of model checking},
  author={Clarke, Edmund M and Henzinger, Thomas A and Veith, Helmut and Bloem, Roderick and others},
  volume={10},
  year={2018},
  publisher={Springer}
}

@book{rubinstein2004CEM,
  title={The cross-entropy method: {A} unified approach to combinatorial optimization, {Monte} {Carlo} simulation, and machine learning},
  author={Rubinstein, Reuven Y and Kroese, Dirk P},
  year={2004},
  publisher={Springer}
}

@article{kim2016improvingAircraftCollisionCEM,
  title={Improving aircraft collision risk estimation using the cross-entropy method},
  author={Kim, Youngjun and Kochenderfer, Mykel J},
  journal={Journal of Air Transportation},
  volume={24},
  number={2},
  pages={55--62},
  year={2016},
  publisher={American Institute of Aeronautics and Astronautics}
}

@article{okelly2018scalableAVTestingCEM,
  title={Scalable end-to-end autonomous vehicle testing via rare-event simulation},
  author={O'Kelly, Matthew and Sinha, Aman and Namkoong, Hongseok and Tedrake, Russ and Duchi, John C},
  journal=neurips,
  volume={31},
  year={2018}
}

@article{norden2019efficient,
  title={Efficient black-box assessment of autonomous vehicle safety},
  author={Norden, Justin and O'Kelly, Matthew and Sinha, Aman},
  journal={arXiv preprint arXiv:1912.03618},
  year={2019}
}

@article{treiber2000IDM,
  title={Congested traffic states in empirical observations and microscopic simulations},
  author={Treiber, Martin and Hennecke, Ansgar and Helbing, Dirk},
  journal={Physical Review E},
  volume={62},
  pages={1805},
  year={2000},
}

@inproceedings{tuncali2019rapidly,
  title={Rapidly-Exploring Random Trees for Testing Automated Vehicles},
  author={Tuncali, Cumhur Erkan and Fainekos, Georgios},
  booktitle=itsc,
  pages={661--666},
  year={2019},
  organization={IEEE}
}

@inproceedings{dreossi2015efficient,
  title={Efficient guiding strategies for testing of temporal properties of hybrid systems},
  author={Dreossi, Tommaso and Dang, Thao and Donz{\'e}, Alexandre and Kapinski, James and Jin, Xiaoqing and Deshmukh, Jyotirmoy V},
  booktitle=nfm,
  year={2015},
}

@InProceedings{Akazaki2018falsification,
author="Akazaki, Takumi and Liu, Shuang and Yamagata, Yoriyuki and Duan, Yihai and Hao, Jianye",
title="Falsification of Cyber-Physical Systems Using Deep Reinforcement Learning",
booktitle=fm,
year="2018",
publisher="Springer International Publishing",
}

@inproceedings{sinha2020neural,
  title={Neural bridge sampling for evaluating safety-critical autonomous systems},
  author={Sinha, Aman and O'Kelly, Matthew and Tedrake, Russ and Duchi, John C},
  booktitle=neurips,
  year={2020}
}

@article{straubinger2020overviewUAM,
  title={An overview of current research and developments in urban air mobility--Setting the scene for {UAM} introduction},
  author={Straubinger, Anna and Rothfeld, Raoul and Shamiyeh, Michael and B{\"u}chter, Kai-Daniel and Kaiser, Jochen and Pl{\"o}tner, Kay Olaf},
  journal={Journal of Air Transport Management},
  volume={87},
  pages={101852},
  year={2020},
}

@article{badue2021selfdrivingsurvey,
  title={Self-driving cars: A survey},
  author={Badue, Claudine and Guidolini, R{\^a}nik and Carneiro, Raphael Vivacqua and Azevedo, Pedro and Cardoso, Vinicius B and Forechi, Avelino and Jesus, Luan and Berriel, Rodrigo and Paixao, Thiago M and Mutz, Filipe and others},
  journal={Expert Systems with Applications},
  volume={165},
  pages={113816},
  year={2021},
}

@article{zhao2017acceleratedIS,
  title={Accelerated evaluation of automated vehicles in car-following maneuvers},
  author={Zhao, Ding and Huang, Xianan and Peng, Huei and Lam, Henry and LeBlanc, David J},
  journal=ieeetits,
  volume={19},
  number={3},
  pages={733--744},
  year={2017},
}

@article{lee2020adaptive,
  title={Adaptive stress testing: Finding likely failure events with reinforcement learning},
  author={Lee, Ritchie and Mengshoel, Ole J and Saksena, Anshu and Gardner, Ryan W and Genin, Daniel and Silbermann, Joshua and Owen, Michael and Kochenderfer, Mykel J},
  journal=jair,
  volume={69},
  pages={1165--1201},
  year={2020}
}

@inproceedings{
  song2020score,
  title={Score-Based Generative Modeling through Stochastic Differential Equations},
  author={Yang Song and Jascha Sohl-Dickstein and Diederik P Kingma and Abhishek Kumar and Stefano Ermon and Ben Poole},
  booktitle=iclr,
  year={2021},
}

@inproceedings{ho2020denoising,
  title={Denoising diffusion probabilistic models},
  author={Ho, Jonathan and Jain, Ajay and Abbeel, Pieter},
  booktitle=neurips,
  volume={33},
  year={2020}
}

@inproceedings{nichol2021improved,
  title={Improved denoising diffusion probabilistic models},
  author={Nichol, Alexander Quinn and Dhariwal, Prafulla},
  booktitle=icml,
  year={2021},
}

@article{gibson2023flowgenmodels,
  title={A Flow-Based Generative Model for Rare-Event Simulation},
  author={Gibson, Lachlan and Hoerger, Marcus and Kroese, Dirk},
  journal={arXiv preprint arXiv:2305.07863},
  year={2023}
}

@article{peltomaki2023requirementgans,
  title={Requirement falsification for cyber-physical systems using generative models},
  author={Peltom{\"a}ki, Jarkko and Porres, Ivan},
  journal={arXiv preprint arXiv:2310.20493},
  year={2023}
}

@inproceedings{janner2022planning,
  title={Planning with Diffusion for Flexible Behavior Synthesis},
  author={Janner, Michael and Du, Yilun and Tenenbaum, Joshua and Levine, Sergey},
  booktitle=icml,
  year={2022},
}

@inproceedings{carvalho2023motion,
  title={Motion planning diffusion: Learning and planning of robot motions with diffusion models},
  author={Carvalho, Joao and Le, An T and Baierl, Mark and Koert, Dorothea and Peters, Jan},
  booktitle=iros,
  year={2023},
}

@article{cerou2007adaptive,
  title={Adaptive multilevel splitting for rare event analysis},
  author={C{\'e}rou, Fr{\'e}d{\'e}ric and Guyader, Arnaud},
  journal={Stochastic Analysis and Applications},
  volume={25},
  number={2},
  pages={417--443},
  year={2007},
  publisher={Taylor \& Francis}
}

@inproceedings{delecki2023model,
  title={Model-based Validation as Probabilistic Inference},
  author={Delecki, Harrison and Corso, Anthony and Kochenderfer, Mykel},
  booktitle={Learning for Dynamics and Control Conference},
  year={2023},
}

@inproceedings{heidlauf2018verification,
  title={Verification Challenges in {F}-16 Ground Collision Avoidance and Other Automated Maneuvers},
  author={Heidlauf, Peter and Collins, Alexander and Bolender, Michael and Bak, Stanley},
  booktitle={ARCH@ ADHS},
  year={2018}
}

@InProceedings{naeem2020fidelitydiversity,
  title = 	 {Reliable Fidelity and Diversity Metrics for Generative Models},
  author =       {Naeem, Muhammad Ferjad and Oh, Seong Joon and Uh, Youngjung and Choi, Yunjey and Yoo, Jaejun},
  booktitle = 	 icml,
  year = 	 {2020},
}

@inproceedings{corso2021transfer,
  title={Transfer learning for efficient iterative safety validation},
  author={Corso, Anthony and Kochenderfer, Mykel J},
  booktitle={Proceedings of the AAAI Conference on Artificial Intelligence},
  volume={35},
  number={8},
  pages={7125--7132},
  year={2021}
}

@inproceedings{baheri2023safety,
  title={Safety validation of learning-based autonomous systems: a multi-fidelity approach},
  author={Baheri, Ali},
  booktitle={Proceedings of the AAAI Conference on Artificial Intelligence},
  volume={37},
  number={13},
  pages={15432--15432},
  year={2023}
}

@inproceedings{zhang2023shiftddpms,
  title={ShiftDDPMs: exploring conditional diffusion models by shifting diffusion trajectories},
  author={Zhang, Zijian and Zhao, Zhou and Yu, Jun and Tian, Qi},
  booktitle={Proceedings of the AAAI Conference on Artificial Intelligence},
  volume={37},
  number={3},
  pages={3552--3560},
  year={2023}
}

@inproceedings{wang2024diffail,
  title={DiffAIL: Diffusion Adversarial Imitation Learning},
  author={Wang, Bingzheng and Wu, Guoqiang and Pang, Teng and Zhang, Yan and Yin, Yilong},
  booktitle={Proceedings of the AAAI Conference on Artificial Intelligence},
  volume={38},
  number={14},
  pages={15447--15455},
  year={2024}
}

@article{mcinnes2018umap,
  title={Umap: Uniform manifold approximation and projection for dimension reduction},
  author={McInnes, Leland and Healy, John and Melville, James},
  journal={arXiv preprint arXiv:1802.03426},
  year={2018}
}

\end{document}